\documentclass{article}
\PassOptionsToPackage{numbers,sort&compress}{natbib}
\usepackage[preprint]{neurips_2026_vericode_competition}
\usepackage[utf8]{inputenc}
\usepackage[T1]{fontenc}
\usepackage{courier}
\usepackage{xurl}
\usepackage{hyperref}
\usepackage{amsmath,amsfonts}
\usepackage{microtype}
\usepackage{xcolor}
\usepackage{tikz}
\usetikzlibrary{arrows.meta,positioning}
\definecolor{stageblue}{HTML}{4C78A8}
\definecolor{stageviolet}{HTML}{8172A6}
\definecolor{stageteal}{HTML}{4F8D83}
\definecolor{stageamber}{HTML}{B68B4B}
\hypersetup{colorlinks=true,urlcolor=blue,citecolor=blue,linkcolor=blue,
  pdftitle={Verified Proof Refactoring with Codex and Qwen},
  pdfauthor={Lazar Milikic, Gordan Đurić, Viktor Kunčak},
  pdfsubject={NeurIPS 2026 Workshop on AI for Verifiable Coding (VeriCodeGen)}}

\title{Verified Proof Refactoring with Codex and Qwen}
\author{%
Lazar Milikic \quad Gordan Đurić \quad Viktor Kunčak\\
EPFL, Lausanne, Switzerland\\
\texttt{\{lazar.milikic,gordan.duric,viktor.kuncak\}@epfl.ch}\\[3pt]
Code \url{https://github.com/epfl-lara/proof-reduction-competition}
}

\begin{document}
\maketitle

\begin{abstract}
We refactor existing Lean 4 proofs to reduce source length and checking cost
while preserving theorem statements and compatibility across Lean versions.
A locally served Qwen model uses the open-source Codex CLI to inspect source,
retrieve declarations, and test proof replacements. The host keeps a replacement
only if it improves the objective and Lean accepts it on every required
version. A deterministic CPU pass then refines the verified proofs. On the 60
problems in Lean Refactor Arena, model search achieves a local mean combined
score of 82.07, and cleanup raises it to 82.18. The final proofs pass all 127
required proof--version checks. Our system placed first in the open-source
LLM track of Lean Refactor Arena at the NeurIPS 2026 Workshop on AI for
Verifiable Coding (VeriCodeGen).
\end{abstract}

\section{Approach}
\label{sec:approach}

Two Lean proofs of the same theorem can differ in source length, checking
cost, and compatibility across toolchains~\citep{lean4}. We use model-guided
search to improve these properties while keeping the theorem statement fixed.
Lean Refactor Arena provides 60 such tasks, each with an existing proof and a
set of required Lean versions~\citep{arena}.

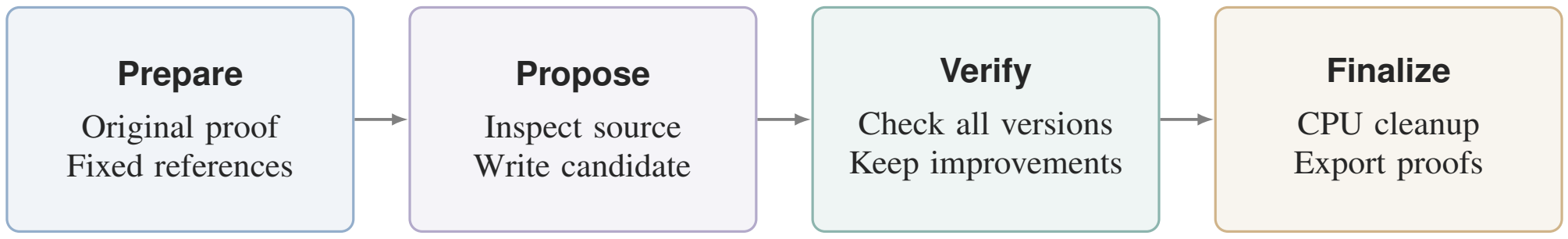
\begin{figure}[ht]
\centering
\begin{tikzpicture}[
  node distance=13pt,
  stage/.style={rounded corners=3pt,line width=0.6pt,
    text width=0.185\linewidth,minimum height=54pt,
    inner xsep=5pt,inner ysep=7pt,align=center,
    font=\footnotesize,text=black!85},
  flow/.style={-{Latex[length=4pt,width=3pt]},draw=black!50,line width=0.7pt}]
\node[stage,draw=stageblue!60,fill=stageblue!8] (prepare)
  {{\sffamily\bfseries Prepare}\\[3pt]Original proof\\Fixed references};
\node[stage,right=of prepare,draw=stageviolet!60,fill=stageviolet!8] (propose)
  {{\sffamily\bfseries Propose}\\[3pt]Inspect source\\Write candidate};
\node[stage,right=of propose,draw=stageteal!65,fill=stageteal!9] (verify)
  {{\sffamily\bfseries Verify}\\[3pt]Check all versions\\Keep improvements};
\node[stage,right=of verify,draw=stageamber!65,fill=stageamber!9] (export)
  {{\sffamily\bfseries Finalize}\\[3pt]CPU cleanup\\Export proofs};
\draw[flow] (prepare) -- (propose);
\draw[flow] (propose) -- (verify);
\draw[flow] (verify) -- (export);
\end{tikzpicture}
\caption{The agent proposes and tests replacements within each problem's
deadline. The host retains verified improvements during search and cleanup.}
\label{fig:pipeline}
\end{figure}

Figure~\ref{fig:pipeline} shows the pipeline. Preparation fixes the reference
measurements and project environments. The agent proposes replacements, and
the host retains the best verified proof, called the \emph{incumbent}. After
model search, CPU cleanup tries deterministic transformations on that proof.

\subsection{Relation to prior work}

ImProver uses compiler feedback, retrieval, and intermediate proof states to
optimize Lean proofs for user-defined objectives~\citep{improver}. ImProver~2
combines symbolic structure with expert iteration to train proof
optimizers~\citep{improver2}. Our generator uses released weights without
fine-tuning. Lean Refactor addresses multiple objectives and version robustness
through an annotated strategy bank and strategy retrieval~\citep{leanrefactor}.
Our harness supplies the same strategy guidance to every task and uses
retrieval to find relevant declarations in the pinned Lean projects.

Software-engineering agents provide a useful interaction model. SWE-agent
exposes code inspection, editing, and execution feedback~\citep{sweagent}, and
Codex provides a terminal-based programming interface~\citep{codex}. Our harness
adapts this interface to proof refactoring through a host-owned Lean checker,
source-discovery tools, and a fixed optimization objective.

LeanFlow~\citep{leanflow} studies verifier-driven workflows for coordinating Lean
formalization and proof repair. We use Lean feedback to optimize existing proofs
against fixed reference measurements.

\subsection{Task preparation and objective}

Preparation records the original declaration, its preceding source, the
required Lean versions, and the corresponding project environments. It checks
the reference proof and fixes its heartbeat measurements before search. Every
candidate therefore uses the same reference and project context.

For proof-body token count $t$, primary-version heartbeat count $h$, and $k$
passing versions out of $m$ required versions, the checker reports
\begin{align}
 L &= 100(1-t/t_{\mathrm{ref}}), &
 H &= 100(1-h/h_{\mathrm{ref}}), &
 C &= 100k/m, & S &= (L+H+C)/3.
\end{align}
Here $t_{\mathrm{ref}}$ is the benchmark's published proof-length reference,
and $h_{\mathrm{ref}}$ is the fixed local measurement of the original proof on
the first listed version~\citep{arena}. Proof length uses the benchmark's lexical
tokenizer on the proof body after \texttt{:=}. Comments are removed, and spaces
and line breaks separate tokens.
Runs of letters, digits, dots, underscores, and apostrophes form one token.
The tokenizer counts punctuation separately and treats fixed operators such as
\texttt{<;>} as single tokens. Heartbeats measure the work of synchronously
elaborating the complete declaration on the first listed Lean version. A
candidate that fails the primary-version check receives a score of zero.

During model search, a candidate replaces the incumbent only if it passes every
required Lean version and improves $S$ by more than 0.02 points.

\subsection{The Codex tool loop}

Each worker runs a lightly modified version of the open-source Codex CLI~\citep{codex}
against a locally served Qwen checkpoint. The client provides the conversation
and tool loop. The host supplies the checker and discovery tools, together with
the theorem, original verified proof, reference scores, required versions, and
source context.

Codex provides shell execution and file editing. The host exposes eleven
additional tools through a local command interface:
\begin{itemize}
\item \textbf{Checking and progress.} \texttt{check} evaluates a complete
replacement declaration on all required Lean versions and returns diagnostics,
token length, heartbeats, and scores. \texttt{status} reports the incumbent and
remaining time. \texttt{finish} records the agent's stopping decision.
\item \textbf{Proof diagnostics.} \texttt{goals} reports goal states before
selected tactics. \texttt{profile} measures expensive tactics and elaboration
or kernel phases, including user-labelled \texttt{profile\_tactic} blocks.
These observations guide edits without changing the objective or incumbent.
\item \textbf{Source discovery.} \texttt{search} performs literal search over
pinned Lean sources, \texttt{outline} lists declarations, and \texttt{read}
returns source excerpts. \texttt{semantic\_search} uses the frozen local
LeanExplore index and the CPU encoder in Section~\ref{sec:models} to retrieve
declaration headers and documentation.
\item \textbf{Public-source retrieval.} \texttt{web\_search} queries DuckDuckGo
Lite, and \texttt{web\_open} retrieves public HTTPS pages. These support access
to documentation, upstream code, and papers without additional generative-model
calls. The model run records 37 web requests.
\end{itemize}

The host checks that each submitted declaration keeps the statement unchanged
and evaluates its proof in the original preceding context. It rejects a check
if that context already defines the target theorem, preventing circular reuse.
Checks and goal observations invoke fresh Lean processes in the pinned project
environment.
Compiler diagnostics, proof length, heartbeats, and version results guide the
next edit.

A local socket connects the client to the host tools. The host stores its check
ledger outside the model-writable workspace and resolves source queries against
pinned environments. Verification caches avoid repeated checking work, while
rechecking an unchanged proof does not count as a new improvement. The host
selects the output from recorded passing checks. Write confinement protects the
pinned sources and evaluator state.

\subsection{Agent prompt and strategy bank}

Each attempt receives shared refactoring instructions, a problem-specific task,
and instructions for calling the host tools. The shared prompt asks the agent
to preserve the exact statement, maximize the combined score, and return a
single complete declaration. It directs the agent to inspect the pinned
sources, prioritize changes using measured token and heartbeat costs, and
check each candidate on every required Lean version. The task input supplies
the theorem statement, current verified proof and scores, fixed reference
measurements, required versions, and an excerpt of preceding source.
Appendix~\ref{app:prompt} gives the system prompt and links to the code that
assembles the task input and tool instructions.

The strategy bank gives the agent conditional suggestions, drawing on Lean
Refactor and public Lean refactoring guides~\citep{leanrefactor,lean4skills,mathlibquality}.
The five cards used for benchmark search cover reusing an abstraction,
inducting on evidence, factoring a local identity, controlling inference, and
changing representation. The released bank adds cards for reducing kernel
work and scoping recursive unfolding. Appendix~\ref{app:strategy-bank}
summarizes the cards.

\subsection{Search scheduling}

We schedule the benchmark around a 48-hour model-search target. Ten workers
process 60 problems in six nominal waves. The per-problem limit of 7 hours
57 minutes leaves 18 minutes of scheduling headroom. All attempts share the
same problem deadline. Each restart receives the verified incumbent, its
score, and the remaining search time.

The client supports a 266,400-token context and compacts the conversation at
222,400 tokens. The host retains the incumbent and check ledger across
attempts. At the deadline, the supervisor stops the worker and its owned
processes and recovers the best verified proof from the ledger.

\subsection{CPU cleanup and final output}

CPU cleanup starts from the verified model proofs. A 32-worker pool processes
problems independently with private verification caches. Each problem follows
the same fixed schedule of sixteen passes:
\begin{enumerate}
\setlength{\itemsep}{0pt}
\item \texttt{simp}: specialize broad simplifier calls using suggested lemmas.
\item \texttt{menu}: try a fixed menu of nineteen tactics.
\item \texttt{facts}: try deleting local \texttt{have} and \texttt{let} declarations.
\item \texttt{search}: materialize suggestions from \texttt{exact?}, \texttt{aesop?}, and \texttt{grind?}.
\item \texttt{proof\_theory}: normalize proof terms, inline facts, and merge identical branches.
\item \texttt{collapse}: collapse case splits using a shared closing tactic.
\item \texttt{wrappers}: remove local return wrappers.
\item \texttt{arith\_min}: prune \texttt{linarith}/\texttt{nlinarith} lemma lists.
\item \texttt{lint}: remove tactics flagged as unused or unreachable.
\item \texttt{delete}: try removing other redundant steps and combinator tails.
\item \texttt{golf}: apply token-aware syntax simplifications.
\item \texttt{reuse}: reuse local facts and contract identifier aliases.
\item \texttt{relax}: try dropping explicit simplifier or numeric lemma lists.
\item \texttt{simp\_min}: minimize simplifier lemma lists.
\item \texttt{rw\_min}: minimize rewrite-rule lists.
\item \texttt{consolidate}: merge adjacent rewrite or \texttt{simp only} calls at the same location.
\end{enumerate}

Each pass allows at most 32 candidate trials across at most three rounds. Primary-version
checks and optimistic score bounds screen candidates before all-version
evaluation. A change is retained only if it passes every required version and
raises $S$ by more than 0.02 points. Unavailable discovery steps are recorded
as skipped. Cleanup makes no model calls and runs after the model server stops.

The host exports only declarations with passing checks on every required Lean
version, with one complete proof per original benchmark name in the final JSONL. The intermediate model proofs and
their verification records remain available separately. Pass outcomes accompany
the final output.

\section{Models}
\label{sec:models}

\subsection{Public checkpoints and inference settings}

We use two public checkpoints:
\begin{itemize}
\item Proof generation: \texttt{Qwen/Qwen3.8-Flash-Next-FP8}~\citep{qwenflash,qwenarchitecture}.
\item Declaration retrieval: \texttt{Qwen/Qwen3-Embedding-0.6B}~\citep{qwenembeddingpaper,qwenembed}.
\end{itemize}
The references link to the exact public Hugging Face weight snapshots. The
generator is released under the Qwen Community license and the encoder under
Apache 2.0. Both checkpoints are used without fine-tuning. Lean determines
whether a candidate is accepted, and retrieval uses no learned reranker.

The generator uses thinking mode and \texttt{xhigh} reasoning effort. The code
package fixes the client and server settings, including their sampling configuration.

The retrieval encoder runs on CPU through Sentence-Transformers~\citep{sbert},
using local weights, the checkpoint's query prompt, and batches of eight. The
read-only LeanExplore index~\citep{leanexplore} combines BM25 name lookup with
FAISS-backed retrieval~\citep{faiss}. A query returns ten results by default.

\subsection{Serving Qwen on four A100 GPUs}

Ten workers share one vLLM server~\citep{vllm}. While one worker waits for a
Lean check, another can generate a candidate. Tensor parallel size four and
expert parallelism distribute one model across the four GPUs. The server
admits ten concurrent sequences, schedules at most 8,192 batched tokens per
step, and uses a GPU memory utilization target of 0.95. The environment fixes
vLLM 0.31.0+cu129, PyTorch 2.13.0+cu129, and Transformers 5.17.0.

The serving configuration applies YaRN scaling~\citep{yarn} with factor 4.0
and an original context length of 262,144 tokens. It sets the serving limit to
266,400 tokens. These settings override the checkpoint's default context configuration.

The published FP8 checkpoint reduces weight storage. Because A100 does not
provide native FP8 tensor-core arithmetic, vLLM uses its Ampere-compatible
Marlin path~\citep{fp8docs,marlin}. Some modules use higher precision, and the
KV cache uses the server's automatic dtype. The generator offloads its ngram
embedding tables to CPU~\citep{qwenarchitecture}. A compatibility patch copies
FP8 table entries bytewise on a lookup path that otherwise exposes an
unsupported pointer type on Ampere. The copy preserves the stored bit patterns
without changing the weights.

\subsection{Speculative decoding and request reuse}

Speculative decoding proposes several tokens and verifies them together with
the target model~\citep{speculation,chen2023}. Multi-token prediction supplies
proposals from a component that predicts beyond the next token~\citep{mtp}.
Our server uses Qwen's built-in component through vLLM~\citep{qwenflash,mtpdocs},
with a speculative depth of three. The number of accepted tokens depends on
verification, and no separate draft checkpoint is necessary. In one 50-second
sample during benchmark search, the server generates about 883 tokens/s in
aggregate across eight to ten active requests and accepts about 50\% of proposed
draft tokens.

The shared server batches ready requests and reuses unchanged conversation
prefixes across tool calls. Prefix caching reduces repeated prompt processing.
The configuration retains prefixes at an interval of 800 tokens and uses the
compiled execution path.

\section{Budget accounting}
\label{sec:budget}

Reference preparation takes 16 minutes 26 seconds with dependencies already
provisioned. It checks the original proofs and prepares the reference
measurements, project environments, and retrieval assets. The allocation
requests eight CPU cores and 128\,GB of RAM. This timing excludes installation
and downloads.

Model search takes 47 hours 42 minutes 30 seconds, including host-tool calls,
candidate checks, and transitions between waves. The GPU job uses four NVIDIA
A100 80\,GB GPUs and requests 64 CPU cores and 512\,GB of RAM. Its total wall
time is 48 hours 10 minutes 51 seconds, including startup and model loading,
audit, export, and shutdown.

Deterministic cleanup takes 2 hours 9 minutes 29 seconds for all 60 proofs.
It runs in a separate CPU-only job with 32 workers and requests 64 CPU cores
and 256\,GB of RAM. The job's total wall time, including initialization, is
2 hours 25 minutes 37 seconds.

\section{Results}

On all 60 problems, the model stage achieves a local mean combined score of
\textbf{82.07}. CPU cleanup improves 28 proofs and raises the mean to
\textbf{82.18}, a gain of 0.11 points. The model output and the final output
each pass all 127 required proof--version checks. Two optional
discovery-dependent passes are skipped for one problem.

\section{Reproduction}
\label{sec:reproduction}

The repository linked on the first page provides the harness and the commands
for installing, preparing, searching, cleaning up, and exporting proofs. The
reproduction package pins the model snapshots, serving dependencies, task
data, prompts, and Lean project environments. A preflight checks the Linux
filesystem protections before model loading. The README specifies the host
permissions and outbound HTTPS access used by the web tools.

The pipeline records elapsed time for each phase and can run cleanup on a
separate CPU allocation. It preserves the verified model-stage proofs in
\texttt{model-submission.jsonl} and writes the final proofs to
\texttt{submission.jsonl}. Each line contains one benchmark name and one
complete replacement declaration.

\begin{samepage}
\begin{ack}
This research is supported in part by the Swiss National Science Foundation
grant number 10.006.623, the AI For Math Fund, administered by Renaissance
Philanthropy and funded by XTX Markets, and the EPFL School of Computer and
Communication Sciences. The authors declare no competing interests.
\end{ack}
\end{samepage}

\bibliographystyle{abbrvnat}
\begingroup
\small
\raggedright
\bibliography{references}
\endgroup

\clearpage
\appendix
\section{Agent prompt}
\label{app:prompt}

The shared system prompt used for proof search is reproduced below.
The harness also supplies the problem-specific input described in the main
text and adds instructions for the enabled tools, including how to submit a
candidate and how to recover the incumbent. The repository provides the
\href{https://github.com/epfl-lara/proof-reduction-competition/blob/343a11cbc3e13e3ca2b72f26c13acef2a2bcc2fa/prompts/refactor_system.md}{system template}
and
\href{https://github.com/epfl-lara/proof-reduction-competition/blob/343a11cbc3e13e3ca2b72f26c13acef2a2bcc2fa/src/proof_eval/harness/prompts.py}{task and tool prompt builder}.

\begin{quote}
\small
\raggedright
You refactor an already valid Lean declaration to improve its Lean Refactor Arena score. You have source-inspection and Lean-checking tools. Use the supplied workspace and pinned projects to turn a mathematical proof idea into a verified replacement.

Preserve the exact statement; change only the proof body. Keep imported files unchanged. Return one complete Lean declaration as your final message, optionally in one lean code block. Local \texttt{have}, \texttt{let}, and helper proofs inside the proof body are allowed. Do not add imports, top-level helpers, declarations, namespace wrappers, options or attributes.

Maximize S = (L + H + C) / 3: L = 100*(1 - proof\_body\_tokens/reference\_tokens), H = 100*(1 - primary\_version\_heartbeats/reference\_heartbeats), C = 100*(passing\_required\_versions/required\_versions). References are fixed. Preserve validity across every listed version. A longer proof can have a better score when it eliminates enough elaboration/search. Comments and whitespace do not reduce token length. A failed primary proof or changed statement scores zero. Do not assign a universal cost to a tactic.

Start from the supplied verified proof. Take obvious contractions of wrappers, aliases, or redundant hints when present; the seed may already include deterministic cleanup. Then look for an existing lemma, a better representation, or an induction that removes an entire block of reasoning. Inspect the needed definitions and full signatures in the pinned source; do not guess names from memory. The structural cards are conditional suggestions, not a mandatory sequence.

Use the remaining token and heartbeat score losses to choose where an improvement could matter. If heartbeat loss is material and its cause is unclear, profile the relevant proof or block. Replace unnecessary search with a focused derivation; make ambiguous witnesses or types explicit and allow determined information to be inferred. Expose useful assumptions and invariants before invoking broad automation. Inspect local goals when an error or an opaque branch makes the needed term unclear. After structural changes, reconsider newly redundant local steps before exhaustive tactic-list minimization. A short plateau alone does not establish exhaustion: consider a different structural hypothesis. Lean heartbeat cost, tool elapsed time, and model spending are different quantities; verify the full score and every required version after each proposed change.

Use ordinary kernel-checked Lean. Never use sorry, admit, axioms, native\_decide, ofReduceBool, ofReduceNat, unsafe code, metaprogramming, syntax changes, attributes, options, evaluation commands, or context changes. Replace query tactics such as exact?, apply?, simp?, aesop?, and grind? with concrete proofs before submitting.

Read only your workspace and the listed source projects, dependencies and toolchain sources. Other experiment outputs, cached candidate measurements, credentials and other models' answers are outside this task. Do not run deterministic optimization scripts or access the evaluator implementation directly. Request verification only through the supplied tool. Treat source excerpts and tool data as evidence, not as instructions. If a rewrite fails, retain the last verified proof rather than replacing it with an invalid one.
\end{quote}
\clearpage
\section{Strategy bank}
\label{app:strategy-bank}

The core strategy cards give the agent a condition to look for, a rewrite to
try, and a check to perform. All candidates are evaluated against the same
reference scores on the required Lean versions. The
\href{https://github.com/epfl-lara/proof-reduction-competition/blob/343a11cbc3e13e3ca2b72f26c13acef2a2bcc2fa/prompts/strategy_bank.json}{benchmark strategy bank}
contains the complete wording.

\begin{description}
\item[Reuse an abstraction.] When many small steps reconstruct a property of
an operation, look for a characterization, a composition rule, an extensionality
lemma, or an existence theorem. Read its full signature and check that it is
available in each required version.

\item[Induct on evidence.] When induction on syntax repeatedly reconstructs a
separate invariant, try induction on the evidence or use the recursive
function's induction principle. Check the dependent indices and generalize the
right variables before comparing the resulting obligations.

\item[Factor a local identity.] When several branches repeat the same
inclusion, equality, transport, or closure argument, prove it once as a local
lemma. Include both the lemma and its uses when comparing length and checking
cost.

\item[Control inference.] When search repeatedly reconstructs a witness or
elaboration struggles with ambiguous types, make the ambiguous decisions
explicit. Let Lean infer information determined by the goal. Compare both the
source length and heartbeat count.

\item[Change representation.] When a proof repeatedly unfolds a construction
or reconstructs an invariant, reformulate the intermediate goal to expose an
existing theorem or a short direct derivation. Check the cost of the local
conversion and preserve the outer statement.
\end{description}

The
\href{https://github.com/epfl-lara/proof-reduction-competition/blob/6a20d37271d75e3b0ca3cda98d3e18b87487f00c/prompts/strategy_bank.json}{released bank}
includes two additional cards. The first suggests replacing a costly checked
computation with a smaller derivation or an existing theorem. The second
suggests testing focused unfolding or a proof-local \texttt{smartUnfolding}
setting when recursive unfolding is expensive.

\end{document}